\pdfoutput=1

\documentclass[11pt]{article}

\usepackage[preprint]{acl}

\usepackage{microtype}
\usepackage{graphicx}
\usepackage{subfigure}
\usepackage{booktabs} %
\usepackage{multirow}
\usepackage{multicol}
\usepackage{makecell}
\usepackage{adjustbox}
\usepackage{tabularx}
\usepackage{caption}  %
\usepackage{xcolor}    %
\usepackage{soul}
\usepackage{enumitem}
\sethlcolor{green!10}
\usepackage{hyperref}

\usepackage{algorithm}
\usepackage[noend]{algorithmic}
\usepackage[algoruled,nofillcomment,algo2e]{algorithm2e}

\usepackage{amsmath}
\usepackage{amssymb}
\usepackage{mathtools}
\usepackage{amsthm}

\usepackage[capitalize,noabbrev]{cleveref}

\usepackage{times}
\usepackage{latexsym}

\usepackage[T1]{fontenc}

\usepackage[utf8]{inputenc}

\usepackage{microtype}

\usepackage{inconsolata}

\usepackage{graphicx}

\theoremstyle{plain}
\newtheorem{theorem}{Theorem}[section]

\theoremstyle{definition}
\newtheorem{definition}[theorem]{Definition}

\theoremstyle{remark}

\newcommand{\myparagraph}[1]{\vspace{1ex}\noindent{\bf #1}}

\usepackage[colorinlistoftodos,textsize=small]{todonotes}
\newcommand{\F}{Fig.}

\newcommand{\T}{Tab.}

\usepackage{xspace}

\newcommand{\pubdata}{$\mathcal{D}_{\text{Public}}$\xspace}
\newcommand{\pridata}{$\mathcal{D}_{\text{Private}}$\xspace}

\newcommand{\teacher}{$\mathcal{M}$\xspace}
\newcommand{\student}{$\mathcal{M}_{\text{S}}$\xspace}

\newcommand{\loss}{\texttt{LOSS}\xspace}
\newcommand{\mink}{\texttt{Min-K\%}\xspace}
\newcommand{\minkpp}{\texttt{Min-K\%++}\xspace}
\newcommand{\zlib}{\texttt{Zlib}\xspace}
\newcommand{\pretrain}{\texttt{Pretrain-Ref}\xspace}
\newcommand{\stable}{\texttt{StableLM}\xspace}
\newcommand{\mope}{\texttt{MoPe}\xspace}

\newcommand{\kd}{\texttt{KD}\xspace}
\newcommand{\seqkd}{\texttt{SeqKD}\xspace}
\newcommand{\imitkd}{\texttt{ImitKD}\xspace}

\title{Membership and Memorization in LLM Knowledge Distillation}

\author{
 \textbf{Ziqi Zhang\textsuperscript{1}},
 \textbf{Ali Shahin Shamsabadi\textsuperscript{2}},
 \textbf{Hanxiao Lu\textsuperscript{3}}, \\
 \textbf{Yifeng Cai\textsuperscript{1}},
 \textbf{Hamed Haddadi\textsuperscript{2,4}},
\\
\\
 \textsuperscript{1}Peking University,
 \textsuperscript{2}Brave Software,
 \textsuperscript{3}Purdue University
 \textsuperscript{4}Imperial College London,
}

\begin{document}
\maketitle
\begin{abstract}
Recent advances in Knowledge Distillation (KD) aim to mitigate the high computational demands of Large Language Models (LLMs) by transferring knowledge from a large ``teacher'' to a smaller ``student'' model. However, students may inherit the teacher's privacy when the teacher is trained on private data. In this work, we systematically \textit{characterize and investigate membership and memorization privacy risks inherent in six LLM KD techniques}.

\looseness=-1 Using instruction-tuning settings that span seven NLP tasks, together with three teacher model families (GPT-2, LLAMA-2, and OPT), and various size student models, we demonstrate that \textit{all existing LLM KD approaches carry membership and memorization privacy risks} from the teacher to its students. However, \textit{the extent of privacy risks varies across different KD techniques}. We systematically analyse how key LLM KD components (KD objective functions, student training data and NLP tasks) impact such privacy risks. 
We also demonstrate \textit{a significant disagreement between memorization and membership} privacy risks of LLM KD techniques. 
Finally, we characterize per-block privacy risk and demonstrate that the privacy risk varies across different blocks by a large margin.    
Our code is available at \url{https://github.com/ziqi-zhang/LLM_Distillation_Privacy}. 
\end{abstract}

\section{Introduction}

\begin{figure*}[th]
\centering
\includegraphics[width=0.9\linewidth]{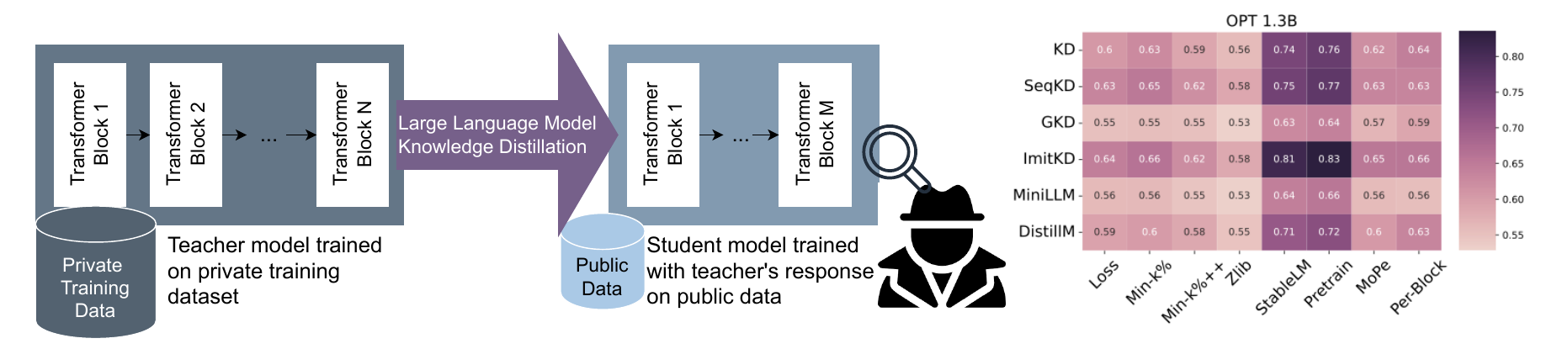}
\caption{Privacy risks of LLM KD. An adversary analyses membership and memorization of a teacher by only looking at its students. We quantify the performance of membership inference attacks in terms of AUC: the higher the value, the more the privacy risk. \textbf{Distilling LLM knowledge reveals private membership information, the extent of this privacy risk varies across different knowledge distillation techniques.}}
\label{fig:threat_model}
\end{figure*}

Knowldeg Distillation (KD)~\cite{hinton2015distilling} techniques have gained widespread adoption in practice\footnote{\url{https://www.bbc.co.uk/news/articles/c9vm1m8wpr9o}} because of their performance and privacy benefits.  
KD reduces high computational costs and memory consumption of machine learning models~\cite{xu2024survey}. KD has been also recently adapted to protect the privacy leakage of LLMs~\cite{xiao2023offsite,tang2022mitigating,mazzone2022repeated,shejwalkar2021membership} based on the assumption that distilling knowledge of a large model (teacher) through a public dataset to a small model (student) can protect the privacy of teacher's training data. This privacy-preserving adaptation of KD has been gaining attraction as it gives the dual advantage of i) efficient students deployed on user devices and ii) better utility than provable protections promised by differentially private techniques.

We \textbf{examine the privacy risk in existing LLM KD techniques} (see Figure~\ref{fig:threat_model})--\texttt{KD}~\cite{hinton2015distilling}, \texttt{SeqKD}~\cite{kim2016sequence}, \texttt{GKD}~\cite{agarwal2024policy}, \texttt{ImitKD}~\cite{lin2020autoregressive}, \texttt{MiniLLM}~\cite{gu2023minillm}, and \texttt{DistiLLM}~\cite{ko2024distillm}. We \textbf{define and quantify the membership and memorization privacy leakage of the teacher's private training data post-distillation}.

We comprehensively analyze whether membership information of the teacher's training data can be inferred from its students: determine if a given data sample is in the teacher's training set (member) or not (nonmember). We use seven Membership Inference Attacks (MIAs)--\minkpp~\cite{zhang2024min}, \mink~\cite{shi2023detecting}, \zlib~\cite{carlini2021extracting}, \loss~\cite{yeom2018privacy}, and two reference-based attacks (\stable~\cite{duan2024membership}, \pretrain~\cite{fu2023practical} and \mope~\citep{li2023mope}. As Figure~\ref{fig:threat_model} (right) shows, the adversary can still recover a large amount of membership information of the private data from the students. For instance, \pretrain recovers private membership information with a 0.83 Area Under the Curve (AUC) from student models obtained through \texttt{ImitKD}. The amount of recovered membership information varies across LLM KD techniques: from 0.64 (\texttt{GKD}) to 0.83 (\texttt{ImitKD}).

We take the first step by measuring if and when teacher training data can be memorized by the student. In particular, we measure how much the student reproduces the teacher's training data verbatim~\cite{carlini2023quantifying}. We find that students can memorize 11.35\% of the same samples that the teacher memorizes. We also find that the student-memorized samples exhibit a different pattern with MIA. MIA is more effective in creative writing, general QA, brainstorming, and open QA than closed QA and classification, while the effectiveness of extraction attacks is vice versa. This finding demonstrates that \textbf{memorization is not membership across KD tasks.}

We further study \textbf{the source of the privacy leakage} and its difference among LLM KD techniques by characterizing key components of KD. We create variants of students isolating the effect of each KD component, including the loss function and student-generated training dataset. We find that using reverse KL loss can mitigate MIA compared to KL. Utilizing student-generated data to stabilize the KD process will increase the MIA performance. 
We also reveal the \textbf{privacy-utility-efficiency trade-off in KD}: decreasing student size can improve privacy protection and efficiency, but harms utility.

Finally, as an step towards designing LLM KD techniques with empirical privacy protections, 
we design a framework measuring a more fine-grained privacy leakage in LLMs.  We first break LLMs into a sequence of transformer blocks. We then analyze each block's privacy leakage by measuring the loss difference due to the model parameter perturbation similar to \mope~\cite{li2023mope}. We demonstrate that \textbf{privacy leakage varies across blocks within the same LLM}. Take GPT2-Large as an example, the AUC of MIAs differs significantly from 0.50 (random guess; 34th block) to over 0.65 (5th block). 
We highlight the following contributions:
\begin{itemize}[leftmargin=*,topsep=0pt,itemsep=0pt]
    \item We define and comprehensively assess the \textbf{membership privacy leakage of LLM KD} using seven MIAs and six LLM KD techniques. 
     \item We define the \textbf{memorization privacy leakage of LLM KD }and present the first empirical study measuring whether training examples memorized by a teacher model remain memorized by the student model after KD.
    \item We present a \textbf{per-block privacy analysis framework} and reveal that privacy leakage varies across blocks.
   
\end{itemize}

\section{Problem formulation}
\label{sec:formulation}

As shown in \F~\ref{fig:threat_model},  we consider training a teacher model \teacher on private data $\mathcal{D}_{\text{Private}}$. $\mathcal{M}$ consists of a set of transformer blocks~\cite{vaswani2017attention} with a huge number of parameters {$\theta = \{\theta^l\}_{l=1}^L$}. Each $\theta^l$ represents parameters of $l$-th block.

\myparagraph{Privacy Risk and Computation Costs.}
An institution needs to solve two issues when it wants to deploy a LLM to users' devices.
The first issue is the privacy risk introduced for $\mathcal{D}_{\text{Private}}$ as $\mathcal{M}$ might memorize and unintentionally leak information about their training data~\cite{carlini2023quantifying,biderman2024emergent}. 
A malicious user can utilize the memorization phenomenon to recover the private information in \pridata.
The second issue is the high computation cost. LLMs require many parameters to perform complex operations and cost computation resources. For example, the state-of-the-art LLMs such as GPT-3 and GPT-4 contain over 100B parameters~\cite{liu2024lightweight}.

\subsection{Knowledge Distillation for LLMs}
\label{sec:KD-description}

Knowledge Distillation (KD) techniques~\cite{xiao2023offsite,xu2024survey,hinton2015distilling,kim2016sequence,agarwal2024policy,lin2020autoregressive,gu2023minillm,ko2024distillm} have been proposed to address above issues--privacy risk and computation costs.
KD techniques do so by distilling the knowledge of $\mathcal{M}$ (as a teacher) to a smaller model (called student \student) through a public dataset (i.e., not using private data $\mathcal{D}_{\text{Private}}$ anymore). In particular, the KD pipeline consist of: i) collecting a public dataset \pubdata; ii) construction the KD dataset $\mathcal{D}_{\text{KD}}$ from \pubdata, iii) defining a KD objective function, $\mathcal{L}_{\text{KD}}$; iv) training \student's parameters with $\mathcal{D}_{\text{KD}}$ and supervision from the teacher:
\begin{equation}
\label{eq:KD-loss}
     \mathbb{E}_{(\mathbf{x},\mathbf{y}) \in \mathcal{D}_{\text{KD}}} \ \mathcal{L}_{\text{KD}}(p(\mathbf{y}| \mathbf{x}), p_{\text{S}}(\mathbf{y} | \mathbf{x}) ),
\end{equation}
where $p(\mathbf{y}| \mathbf{x})$ and $p_{\text{S}}(\mathbf{y} | \mathbf{x}) $ are teacher's and student's distribution, and $\mathbb{E}$ is the expectation over the KD dataset.

Researchers have designed various algorithms to distill knowledge more efficiently and steadily, depending on the choice of $\mathcal{D}_{\text{KD}}$ and $\mathcal{L}_{\text{KD}}$. \T~\ref{tab:KD} summarizes six LLM KD techniques: \texttt{KD}~\cite{hinton2015distilling}, \texttt{SeqKD}~\cite{kim2016sequence}, \texttt{GKD}~\cite{agarwal2024policy}, \texttt{ImitKD}~\cite{lin2020autoregressive}, \texttt{MiniLLM}~\cite{gu2023minillm}, and \texttt{DistiLLM}~\cite{ko2024distillm}.  
\texttt{KD} uses Kullback-Leibler (KL) divergence to compute the output distribution difference between \teacher and \student.
\texttt{SeqKD} uses KL loss to train \student with data generated from the teacher $\mathcal{D}_{\text{T}}$. $\mathcal{D}_{\text{T}}=\{ (\mathbf{x}, \mathcal{M}(\mathbf{x})) \}$ is generated by feeding $\mathbf{x} \in \mathcal{D}_{\text{Public}}$ to \teacher and collect teacher's feedback $\mathcal{M}(\mathbf{x})$.
\texttt{ImitKD} uses student feedback dataset $\mathcal{D}_{\text{S}}$ to compute KL divergence. $\mathcal{D}_{\text{S}}$ is constructed by dynamically prompting the under-training student with $\mathbf{x}$.
\texttt{GKD} utilizes the generalized Jensen-Shannon divergence~\cite{menendez1997jensen} loss to train \student. It directly uses \pubdata as $\mathcal{D}_{\text{KD}}$.
\texttt{MiniLLM} proposes the Reverse KL (RKL) function to prevent \student from overestimating the low-probability regions of \teacher's distribution. \texttt{MiniLLM} mixes $\mathcal{D}_{\text{S}}$ with \pubdata to stabilize the training.
\texttt{DistiLLM} proposes a Skewed RKL divergence loss and adaptively mix $\mathcal{D}_{\text{S}}$ with \pubdata to enhance efficiency. The mixture ratio is dynamically adjusted.
Note that both $\mathcal{D}_{\text{T}}$ and $\mathcal{D}_{\text{S}}$ are constructed based on \pubdata and do not have overlap with \pridata.

\begin{table}[!t]
\caption{An overview of LLM KD techniques, highlighting differences in the KD dataset $\mathcal{D}_{\text{KD}}$ and objective function $\mathcal{L}_{\text{KD}}$.}

\label{tab:KD}
\vspace{-6pt}
\setlength{\tabcolsep}{1.5pt}
\centering
\begin{adjustbox}{max width=\linewidth}

\begin{tabular}{@{}lcc@{}}
\toprule
Technique      & $\mathcal{D}_{\text{KD}}$                                                                                                                                                                 &  $\mathcal{L}_{\text{KD}}$                                                                                                                                                                                                                                                                                               \\ \toprule
\texttt{KD}       & \begin{tabular}[c]{@{}c@{}}Public dataset\\ $\mathcal{D}_{\text{Public}}=\{ (\mathbf{x},\mathbf{y}) \}$\end{tabular}                                                                                                                                                         & \begin{tabular}[c]{@{}c@{}}$\text{KL Divergence}$\\ $\mathbb{E}_{\mathbf{x}} \mathbb{E}_{\mathbf{y} \sim p(\cdot \mid \mathbf{x})} \left[ \log \frac{p(\mathbf{y} \mid \mathbf{x})}{p_{\text{S}}(\mathbf{y} \mid \mathbf{x})} \right]$\end{tabular} \\ \midrule
\texttt{SeqKD}    & \begin{tabular}[c]{@{}c@{}}Teacher Feedback\\ $\mathcal{D}_{\text{T}}=\{ (\mathbf{x}, \mathcal{M}(\mathbf{x})) \}$\end{tabular}                                                               & \begin{tabular}[c]{@{}c@{}}$\text{KL Divergence}$\\ $\mathbb{E}_{\mathbf{x}} \mathbb{E}_{\mathcal{M}(\mathbf{x}) \sim p(\cdot \mid \mathbf{x})} \left[ \log \frac{p(\mathcal{M}(\mathbf{x}) \mid \mathbf{\mathbf{x}})}{p_{\text{S}}(\mathcal{M}(\mathbf{x}) \mid \mathbf{x})} \right]$\end{tabular}                                                                                                                                                                                                                                                                                        \\ \midrule
\texttt{ImitKD}   & \begin{tabular}[c]{@{}c@{}}Student Feedback  \\ $\mathcal{D}_{\text{S}}=\{ (\mathbf{x}, \mathcal{M}_S(\mathbf{x})) \}$\end{tabular}                                                             & \begin{tabular}[c]{@{}c@{}}$\text{KL Divergence}$\\ $\mathbb{E}_{\mathbf{x}} \mathbb{E}_{\mathcal{M}_S(\mathbf{x}) \sim p(\cdot \mid \mathbf{x})} \left[ \log \frac{p(\mathcal{M}_S(\mathbf{x}) \mid \mathbf{\mathbf{x}})}{p_{\text{S}}(\mathcal{M}_S(\mathbf{x}) \mid \mathbf{x})} \right]$\end{tabular}                                                                                                                                                                                                                                                                                 \\ \midrule
\texttt{GKD}      & $\mathcal{D}_{\text{Public}}$                                                                                                                                                          & \begin{tabular}[c]{@{}c@{}}$\text{Jensen-Shannon}$\\ $\beta \text{KL}(p, p_{\text{S}}) + (1-\beta) \text{KL}(p_{\text{S}}, p)$\end{tabular}                                                                                                          \\ \midrule
\texttt{MiniLLM}  & \begin{tabular}[c]{@{}c@{}}Mixed Dataset \\ $ \mathcal{D}_{\text{Public}} \cup \mathcal{D}_{\text{S}}$\end{tabular}                                                        & \begin{tabular}[c]{@{}c@{}}Reverse KL\\ $\text{KL}(p_{\text{S}}, p)$\end{tabular}                                                                                                                                                          \\ \midrule
\texttt{DistiLLM} & \begin{tabular}[c]{@{}c@{}}Adaptive Mixed Dataset\\ $ \mathcal{D}_{\text{Public}} \cup \mathcal{D}_{\text{S}}$\end{tabular} & \begin{tabular}[c]{@{}c@{}}Skewed Reverse KL \\ $\text{KL}(p_{\text{S}}, \alpha p + (1-\alpha) p_{\text{S}})$\end{tabular}                                                                                                               \\ \bottomrule
\end{tabular}

\end{adjustbox}
\end{table}

\subsection{Our goal} 
As \student is trained on public data and is not directly trained on \pridata, \student is usually regarded to not contain privacy information in \teacher~\cite{xiao2023offsite,tang2022mitigating,mazzone2022repeated,shejwalkar2021membership}. Our objective is to quantify the privacy protection effect of existing KD techniques. 
We select all six state-of-the-art KD techniques. We reuse the public code and hyper-parameters of \texttt{DistiLLM}~\cite{ko2024distillm} to train the models.

\subsection{Defining Privacy Protection of LLM KD} 

We comprehensively analyze the privacy protection of KD techniques on LLMs concerning two types of privacy attacks: \textbf{Membership inference attack} and \textbf{Data extraction attack}. 

\myparagraph{Related work.} The only related work to ours is \cite{jagielski2024students}, which evaluates a single membership inference attack on a single KD technique (\texttt{KD}) using small ML models. In contrast, we comprehensively study the privacy protection of all six existing KD techniques on LLMs using seven state-of-the-art membership inference attacks. Furthermore, we define and investigate KD memorization through data reconstruction attacks.

\subsubsection{Membership}

MIAs aim to infer whether a specific data record was included in the training dataset of a target model~\cite{shokri2017membership,mattern2023membership,mireshghallah2022empirical,mitchell2023detectgpt,yeom2018privacy,carlini2021extracting,shi2023detecting,zhang2024min,carlini2021extracting,fu2023practical,li2023mope}. 
We define the \textbf{membership inference leakage of LLM KD} as the success of MIAs in inferring the membership of teacher's training data through its student:
\begin{definition}[Knowledge Distilled Membership Privacy Risk]
    Let \teacher be a teacher trained on \pridata, and \student be the student trained using a specific KD technique from \teacher. Given a data point $\mathbf{x}$, we define that the KD inherits a membership privacy risk if there exists a Membership Inference Attack (MIA) that can correctly infer $\mathbf{x}$'s membership status in \pridata by querying \student.  
\end{definition}

We use seven state-of-the-art MIAs encompassing reference-based, black-box, and white-box approaches: \loss~\cite{yeom2018privacy}, \zlib~\cite{carlini2021extracting}, \mink~\cite{shi2023detecting}, \minkpp~\cite{zhang2024min}, \stable~\cite{duan2024membership}, \pretrain~\cite{fu2023practical} and \mope~\cite{li2023mope}. 
\loss computes the cross-entropy loss value to evaluate membership. The core intuition is that training data (members) generally have lower loss values than non-training data (non-members). 
\zlib computes the ratio between per-sample perplexity value and Zlib text entropy for membership inference. 
\mink is based on the hypothesis that an unseen data point will likely contain a few outlier words with low probabilities under the LLM. This algorithm selects the K\% tokens with the lowest confidence and computes the average confidence of these tokens.
\minkpp is based on the insight that training samples tend to be local maxima of the modeled distribution. So, the probabilities should be computed based on the conditional categorical distribution. 
Reference-based attacks consider each target sample's intrinsic complexity and use the loss value on a reference model to calibrate. We use two types of reference-based attacks. \stable follows the latest empirical study~\cite{duan2024membership} and use the best StableLM-Base-Alpha-3B-V2 as the reference model.
\pretrain uses the pre-trained teacher model as the reference model~\cite{fu2023practical}.
\mope is the only white-box MIA. It perturbs the model parameters and uses the model output variance as the metric. The insight is that member data should have a larger loss variance than nonmember data. White-box MIAs are also practical in LLM KD  when the client can access the model weights deployed on their device.

\subsubsection{Memorization}

Data extraction attacks aim to recover individual training data records from a model. LLM memorization is usually defined as $K$-extractible~\cite{carlini2021extracting}, a sample $\mathbf{x}$ is said to be $K$-extractible if it (a) exists in the training dataset, and (b) can be generated by prompting the model with $K$ prior tokens.
We define the \textbf{memorization risk of LLM KD} as the success of attacks in extracting training data of the teacher from its students:
\begin{definition}[Knowledge Distilled Memorization Risk]
Let \teacher be the teacher trained on \pridata, and \student be the student trained using a KD technique from \teacher. Let $\mathbf{x}$ be an example from $\mathcal{D}_{\text{Private}}$, and $\mathbf{x}$ can be split into a prompt $\mathbf{x}_p$ and a victim $\mathbf{x}_v$: $\mathbf{x} = [ \mathbf{x}_p || \mathbf{x}_v ]$. We define that the KD inherits memorization if both \teacher and \student produce $\mathbf{x}_v$ when prompted by $\mathbf{x}_p$: $\mathcal{M}(\mathbf{x}_p)=\mathbf{x}_v \& \mathcal{M}_S(\mathbf{x}_p)=\mathbf{x}_v$.
\end{definition}

\begin{table*}[t]
    \small
    \centering
    \caption{Membership privacy leakage of teachers (GPT2-XL, OPT-2.7B, and LLAMA2-7B) evaluated using the performance (AUC and TPR at various FPRs) of seven MIAs once performing attack directly on teachers directly. Main takeaways: \textbf{i) All teachers exhibit significant membership privacy leakage, though the extend varies across families; and ii) The most successful MIA differs across families and metrics, highlighting the absence of a universally optimal MIA strategy.}}
\label{tab:MIA_teachersGPTOPT}
\begin{tabular}{@{}llccccccc@{}}
\toprule
             &     &   \loss & \mink & \minkpp & \stable & \pretrain & \zlib & \mope \\ \midrule
\multirow{3}{*}{GPT-2 XL}      &    AUC   & 0.9715 & 0.9735 & 0.9371 & 0.9824 & 0.9175 & 0.9774 & 0.6096\\
                                & TPR@05  & 0.8854 & 0.9032 & 0.3487 & 0.8997 & 0.2458 & 0.9703 & 0.0440\\
                                & TPR@01  & 0.1778 & 0.1955 & 0.0609 & 0.6230 & 0.0284 & 0.1998 & 0.0020\\

                                \midrule
\multirow{3}{*}{OPT 2.7B}     &    AUC    & 0.9432 & 0.9532 & 0.9204 & 0.9604 & 0.9806 & 0.8633 & 0.9196\\
                                & TPR@05  & 0.8102 & 0.8652 & 0.2468 & 0.8895 & 0.8703 & 0.7254 & 0.5742\\
                                & TPR@01  & 0.3944 & 0.2162 & 0.0629 & 0.6932 & 0.6420 & 0.3674 & 0.2438\\
                                \midrule
\multirow{3}{*}{LLAMA2-7B}     &    AUC   & 0.8827 & 0.9133 & 0.8836 & 0.8788 & 0.8949 & 0.9293 & 0.7507 \\
                                & TPR@05  & 0.6613 & 0.7655 & 0.5475 & 0.7119 & 0.7544 & 0.7997 & 0.5128\\
                                & TPR@01  & 0.3425 & 0.3260 & 0.0219 & 0.6016 & 0.6494 & 0.7103 & 0.2764\\

\bottomrule
\end{tabular}
\end{table*}

\section{Experimental Setup}

\myparagraph{Dataset}. Following recent LLM KD literature~\cite{ko2024distillm,gu2023minillm}, we consider the instruction-following task using \textit{databricks-dolly-15k}~\cite{DatabricksBlog2023DollyV2} (an open source dataset of instruction-following records generated by thousands of Databricks employees in eight tasks: brainstorming, classification, closed QA, generation, information extraction, open QA, and summarization).  
We follow prior literature~\cite{ko2024distillm} to split the dataset: randomly select 11K samples for training, 1K for validation, and 0.5K for testing. We then evenly divide the training dataset to construct the teacher and student dataset following~\cite{jagielski2024students}. We split the 11K training samples into a teacher set of 5.5K (\pridata) and a student set of 5.5K (\pubdata). We ensured there was no duplication, distiribution shift and $n$-gram similarities between the teacher and the student set (See Appendix~\ref{app:publicvsprivate}). We randomly select 1K samples from the teacher training dataset as members and use 1K validation samples as non-members.

\myparagraph{Teacher/Student LLMs}. We consider three families of LLMs: GPT-2~\cite{radford2019language}, OPT~\cite{zhang2022opt}, and LLAMA-2~\cite{openlm2023openllama}. Following DistiLLM~\cite{ko2024distillm}, i) for the GPT-2 family, we use the GPT-2 XL (1.5B) as the teacher model and GPT-2 Small (124M), GPT-2 Medium (355M), and GPT-2 Large (774M) as the students; ii) for the OPT family, we use OPT-2.7B as the teacher model and OPT-1.3B, OPT-0.3B, and OPT-0.1B as the students; and iii) for the LLAMA family, we use LLAMA2-7B as teacher and LLAMA2-3B as the student.

\myparagraph{Metrics.} We measure the performance of MIAs using four standard metrics~\cite{carlini2022membership,li2023mope,wang2024pandora}: Area Under the Curve (AUC), True Positive Rate (TPR) at low False Positive Rates (FPR) of 5\% and 1\% denoted as TPR@05 and TPR@01, and a log-scale Receiver Operating Characteristic (ROC). The higher the privacy leakage, the higher the AUC, TPR@05, or TPR@01. We also measure memorized tokens. We report the utility of students in Appendix~\ref{app:utility} and check the consistency with~\cite{ko2024distillm}.

\myparagraph{Statistical Significance.}
We select the OPT-1.3B experiments and run them five times to check the statistical significance. The variances of AUC, TPR@05, and TPR@01 are $4e^{-5}$, $6e^{-5}$, and $1e^{-5}$.

\section{Empirical Evaluation}

\subsection{Membership Privacy Leakage of Teacher}

We first analyze membership privacy leakage of teacher models about their private training data. Table~\ref{tab:MIA_teachersGPTOPT} reports the privacy leakage of GPT2-XL, OPT-2.7B, and LLAMA2-7B measured by AUC and TPR at low FPRs of seven MIAs.

\myparagraph{High membership privacy leakage varies across different families of teachers.} All teacher models exhibit significant membership privacy leakage, but the extent of leakage varies. For example, the AUC (averaged across all MIAs) for GPT2-XL, OPT-2.7B, and LLAMA2-7B is 0.9649, 0.9210, and 0.9022, respectively. These differences arise from a combination of the model's inherent leakage and the effectiveness of the attacks. The actual leakage correlates with the model's generalization ability: better generalization leads to reduced memorization. Several factors influence generalization, including i) model size, ii) training set size, iii) number of training iterations, and iv) model architecture.  Our findings emphasize the importance of considering these factors jointly (not in isolation) when evaluating privacy leakage in different models.
Larger models are typically more prone to overfitting training data~\cite{yuan2022membership}. This suggests that MIA performance would be higher for LLAMA2-7B compared to GPT2-XL and OPT-2.7B. However, this is not observed in practice. LLAMA2-7B achieves better generalization, likely due to its significantly larger training dataset~\cite{hoffmann2022training,muennighoff2023scaling} (GPT2-XL, OPT-2.7B, and LLAMA2-7B are trained on 100M, 180B, and 1.4T tokens, respectively), which reduces memorization of specific samples.

\myparagraph{Lack of a single dominant MIA.} No single MIAs consistently outperforms the others across all metrics and teachers. For GPT-2 XL, the highest AUC belongs to \stable, and the highest TPR@05 belongs to \zlib. 
In general, both reference-based attacks (\stable and \pretrain) achieve the highest TPR in the low FPR region. \stable achieves an average TPR@01 and TPR@001 of 0.6393 and 0.1139, respectively, outperforming single-model MIAs by over 2.7708$\times$ and 3.0536$\times$. However, \pretrain performs poorly on GPT2-XL, likely due to the relatively weak reference model (1.5B) compared to other models (over 2.7B). The best-performing MIA varies depending on the metric and model, reflecting the lack of a universally optimal attack. Therefore, \textbf{understanding real privacy leakages of a model requires conducting multiple MIAs across diverse metrics}. 

White-box \mope performs poorly on GPT2-XL. The AUC of \mope is 0.6096, which is 32\% lower than \loss (AUC is 0.9715). This is potentially due to \mope's sensitivity to the hyper-parameter settings and the size of the target model, as noted in the original \mope paper~\cite{li2023mope}. We followed the original paper to set the hyper-parameters, but due to different models and datasets, these hyper parameters may not be optimal for new settings. This observation opens an avenue for future research into \textbf{improving white-box MIAs by leveraging available information to be competitive with or outperform black-box MIAs}.

\begin{figure*}[t]
\centering
\includegraphics[width=1\linewidth]{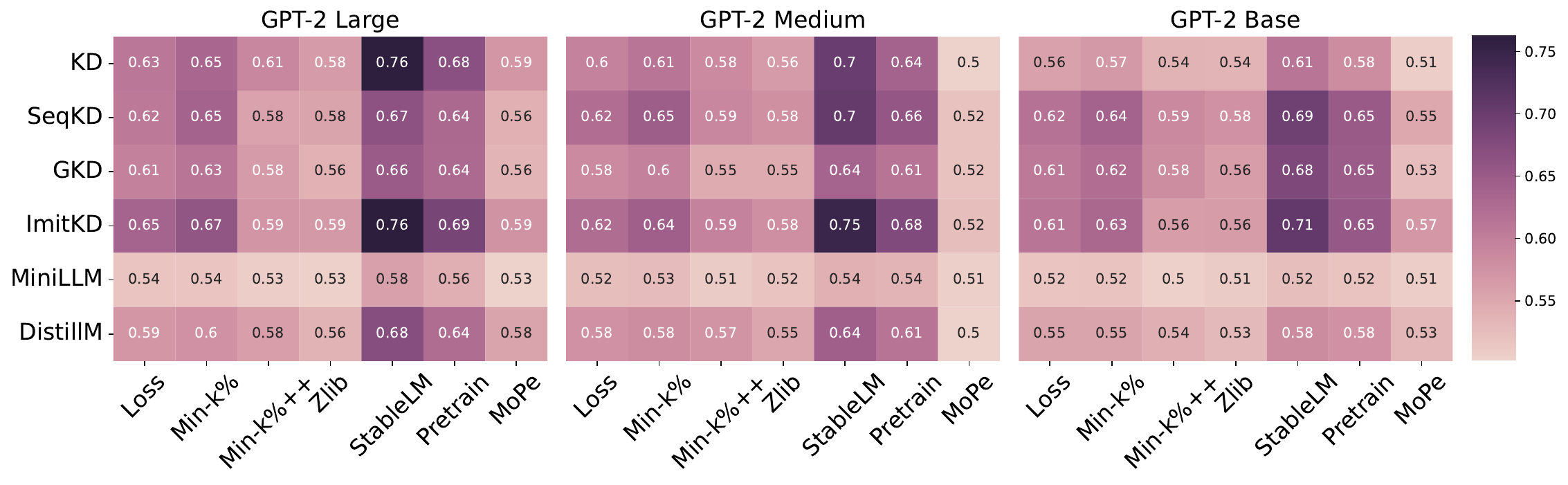}
\includegraphics[width=1\linewidth]{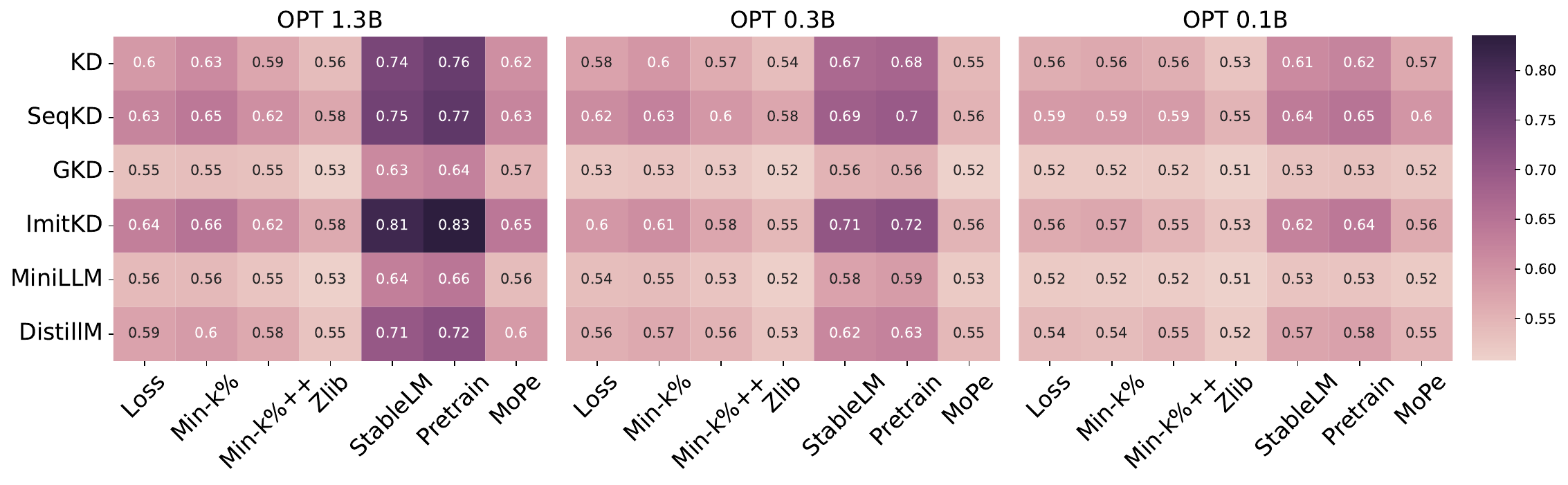}

\caption{Membership privacy protection of six knowledge distillation techniques using GPT-2 (top row) and OPT (bottom row) family of teachers. We report the AUC score of seven MIAs at inferring private training members of teachers based on three students (GPT-2 Large, GPT Medium and GPT Small) and (OPT 1.3B, 0.3B and 0.1B) obtained through various knowledge distillation techniques. \textbf{None of knowledge distillation techniques can create students that protect privacy of their GPT-2 XL teacher.} See Appendix~\ref{app:roc} and Appendix~\ref{app:tpr} for results of ROC curves and TPR at low FPRs.}
\label{fig:model_AUC_GPT2_OPT_studnets}
\vspace{-3ex}
\end{figure*}

\subsection{KD Membership Privacy Risk}

\myparagraph{None of the knowledge distillation techniques can protect the privacy of \teacher's private training data.} Fig.~\ref{fig:model_AUC_GPT2_OPT_studnets} shows the performance of MIAs on students of GPT-2 and OPT (See Appendix~\ref{app:llama} for results on LLAMA2-7B). MIAs can still infer the membership of teacher's training data by only having access to their students. For example, the AUC against \kd on GPT2 models can be over 0.70 (\kd on GPT2-Large in Fig.~\ref{fig:model_AUC_GPT2_OPT_studnets}). For OPT models, the AUC can be over 0.80 (\stable and \pretrain on OPT-1.3B). From Fig.~\ref{fig:model_AUC_GPT2_OPT_studnets}, we can observe that no KD techniques can achieve an AUC under 0.60 and 0.64 against all MIAs on GPT2-Large and OPT-1.3B, respectively. 
TPRs on low FPR regions (see Appendix~\ref{app:tpr}) also demonstrate that MIA on \student can still reveal non-trivial private information. 
For the GPT models, the averaged TPR@05 of GPT2-Large, Medium, and Small are 0.0735, 0.0672, and 0.0638, respectively. Averagely, the TPR@05 is higher than the random-guess baseline (0.05) by 36.33\%. Meanwhile, the averaged TPR@01 is higher by 55.33\%. 
Similarly, for the OPT models, the averaged TPR@05 and TPR@01 are higher than random-guess by 58.80\% and 83.67\%, respectively.
Compared with teachers, the leakage from \student is also non-trivial. The AUC of \stable against GPT-2 Large is an average of 69\% of that for the teacher. For OPT 1.3B, the average AUC of \stable is 71.33\% of the teacher.

\myparagraph{Membership privacy risk varies across different KD techniques.} \kd, \seqkd, and \imitkd have higher AUC than other techniques. On OPT models, the AUCs of \kd, \seqkd, and \imitkd are higher than other solutions by 8.26\%, 12.70\%, and 12.11\%, respectively. \seqkd is more vulnerable at low FPR region. The row of \seqkd is much darker than other rows. On OPT models, \seqkd leads to higher TPR@05 and TPR@01 than other KD by 31.89\% and 66.91\%, respectively. 
On LLAMA, \seqkd also results in an average of 57.72\% higher TPR@05 than other solutions. 
The reason for \seqkd's high privacy leakage is that \seqkd uses \teacher's verbatim output to build $\mathcal{D}_{KD}$. According to Tab.~\ref{tab:MIA_teachersGPTOPT}, \teacher has a high MIA score and thus tends to remember the labels in \pridata. Thus, \teacher's verbatim output is likely to contain sentences in \pridata, and such sentences are included in $\mathcal{D}_{\text{KD}}$. As \student is directly trained on data from \pridata, it memorizes private samples easier, thus high MIA performance.

\myparagraph{The size of the student model affects privacy leakage.} By comparing different columns in Fig.~\ref{fig:model_AUC_GPT2_OPT_studnets}, we can observe that a smaller student model often yields a lower attack performance. For GPT2, the average AUC of GPT2-Large, GPT2-Medium, and GPT2-base are 0.6102, 0.5867, and 0.5764, respectively. The GPT2-base has a 5.54\% lower AUC than GPT2-Large. This observation is also valid for OPT and other metrics. For the AUC of OPT models, OPT-0.3B and OPT-0.1B have a lower AUC compared to OPT-1.3B by 6.98\% and 11.19\%, respectively. For TPR@05 and TPR@01, GPT2-base has a lower value than GPT2-Large by 13.15\% and 18.23\%, respectively. Similarly, OPT-0.1B has a lower TPR@05 and TPR@01 than OPT-1.3B by 27.19\% and 28.09\%, respectively. The reason for the low performance is the limited model capacity. A small model has fewer parameters and thus memorizes less membership information of \teacher. This can be validated by the inferior performance of the smaller student model on the downstream tasks. For GPT2, the smaller models have a lower utility performance than GPT2-Large by 31\%. The smaller OPT models also have a 23\% lower performance.

\myparagraph{The success of estimating privacy leakage of KD varies across different MIAs.} We can observe that the reference-based attack outperforms the single-model black-box attack. In GPT2 and OPT models, \stable consistently achieves the highest AUC, TPR@05, and TPR@01 across all KD techniques. Averagely, \stable achieves 12.47\% higher AUC, 44.82\% higher TPR@05, and 60.50\% higher TPR@01 than other MIAs. This observation is consistent with~\cite{duan2024membership}. However, on the LLAMA model (Figure~\ref{fig:model_AUC_LLAMA_studnets} in Appendix~\ref{app:llama}), \pretrain slightly outperforms \stable. We think the reason is that the reference models of \pretrain and \stable have the same size, but the reference model of \pretrain is more similar to the student model. Thus, \pretrain has a slightly better calibration effect.

\section{Ablation study}
We present an ablation study to understand the impact of design choices in LLM KD techniques. For this study, we report the performance of the most successful attacks \stable and \pretrain.

\begin{figure}[t]
\centering
\includegraphics[width=\linewidth]{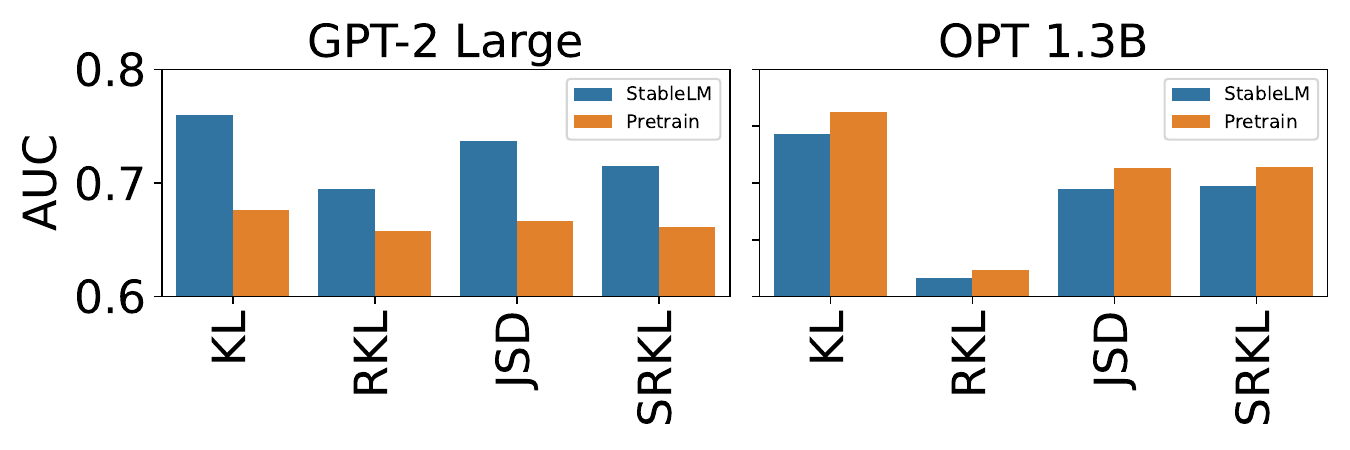}
\caption{The impact of KD loss function on membership privacy protection. We choose \stable and \pretrain because they are the most successful attacks. \textbf{KL has the highest MIA AUC, and RKL can significantly reduce AUC.}}
\vspace{-5pt}
\label{fig:Ablation_Loss_MIA}
\end{figure} 

\myparagraph{Impact of KD Objective Function.}
We start from \kd and change the loss types to train different student models. Then, we evaluate MIA on each model. Figure~\ref{fig:Ablation_Loss_MIA} reports the AUC scores. We can observe that KL leads to the highest MIA. For example, using \stable yields the highest AUC on GPT2-Large (0.76), and using \pretrain achieves 0.74 on OPT-1.3B. On the contrary, RKL has the lowest AUCs for both models. The AUC of JSD is between KL and RKL because JSD is the weighted average of KL and RKL. SRKL increases the AUC compared with RKL.

\begin{figure}[t]
\centering
\includegraphics[width=\linewidth]{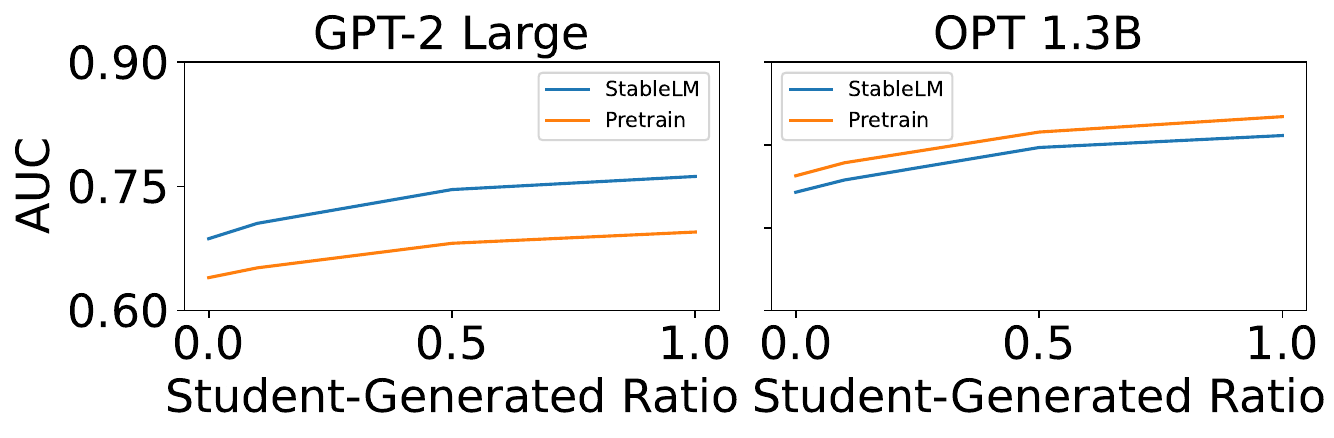}
\vspace{-10pt}
\caption{Ablation study on the ratio of student-generated data. We choose two most successful attacks to perform the study. \textbf{Privacy risk increases as the ratio of student-generated data increases.}}
\vspace{-10pt}
\label{fig:Ablation_Alpha_MIA}
\end{figure} 

\myparagraph{Impact of the Size of Student Feedback.}
\texttt{MiniLLM} and \texttt{DistiLLM} dynamically add different ratios of student-generated output (SGO) to stabilize KD. 
To study the influence of SGO, we fuse SGO data with \pubdata to construct $\mathcal{D}_{\text{KD}}$. We control the ratio of SGO in $\mathcal{D}_{\text{KD}}$ from 0 to 1. 
Figure~\ref{fig:Ablation_Alpha_MIA} shows the performance of MIA w.r.t different ratios. We can observe that as the ratio of SGO increases, the MIA performance increases. For example, on GPT2-Large, the AUC of \stable increases from 0.69 to 0.76. On OPT-1.3B, the AUC of \pretrain increases from 0.76 to 0.83. 
The reason might be that the SGO data contains some data in \pridata with low loss values, and utilizing SGO in training further decreases the loss value and increases MIA.

\myparagraph{Privacy-Utility-Efficiency Trade-Off.}
We also study the relationship between privacy, utility, and efficiency. We compute the relative utility for each student model compared to the teacher and calculate the average score. The relative utility of OPT-1.3B, OPT-0.3B, and OPT-0.1B are 98.24\%, 93.13\%, and 83.34\%, respectively. GPT2-Large, GPT2-Medium, and GPT2-Base have relative utilities of 85.86\%, 82.68\%, and 78.15\%. From Figure~\ref{fig:model_AUC_GPT2_OPT_studnets}, we can observe that as model size decreases, the AUC score decreases. Appendix~\ref{app:tpr} shows a similar observation for TPR@05 and TPR@01. This observation reveals a trade-off between privacy, utility, and efficiency: \textbf{decreasing the model size in KD can improve the on-device efficiency and reduce privacy leakage but harms model utility}.

\section{Memorization of KD}
\label{app:memorization}

\label{sec:data_reconstruction}
\begin{figure}[t]
\centering
\includegraphics[width=1\linewidth]{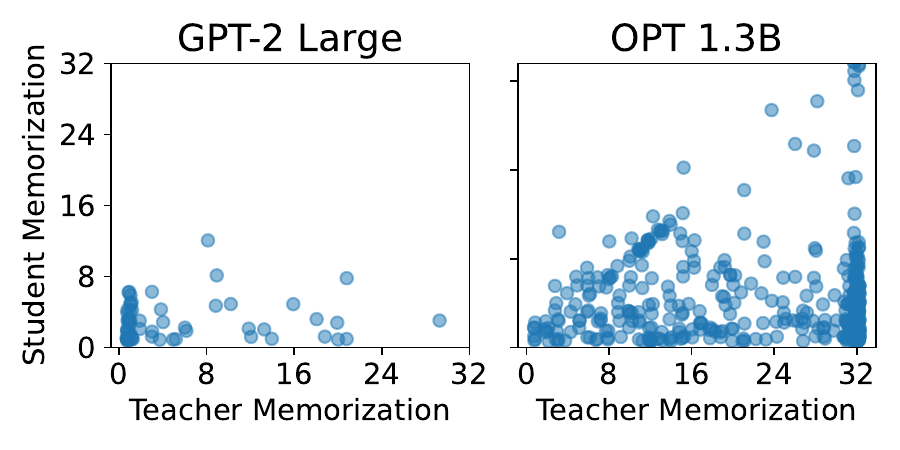}
\vspace{-12pt}
\caption{Data reconstruction attack against the student model on different samples that are memorized by the teacher (absolute value). GPT2-Large has smaller samples because GPT2 teacher model memorizes smaller number of tokens than OPT. 
}

\label{fig:Data_Reconstruction}
\end{figure}

We also study the performance of revealing \pridata's verbatim data from \student, i.e., the relationship between \student's memorized \pridata samples and \teacher's memorized samples. Figure~\ref{fig:Data_Reconstruction} shows how much the teacher or student remembers each sample. Each point represents a data sample. The x-axis and y-axis represent how many tokens the \teacher and \student remember, respectively. We only report the 32-token memorization following~\cite{biderman2024emergent}.
We can observe that \student can remember a non-trivial number of tokens that the teacher memorizes. The adversary can recover part of the tokens from \student output. Note that 11.35\% samples lie on the diagonal of the figure, which means that for 11.35\% samples, \student memorizes the same number of tokens as \teacher. 

\subsection{Memorization Versus Membership}

\begin{figure}[t]
\centering
\includegraphics[width=\linewidth]{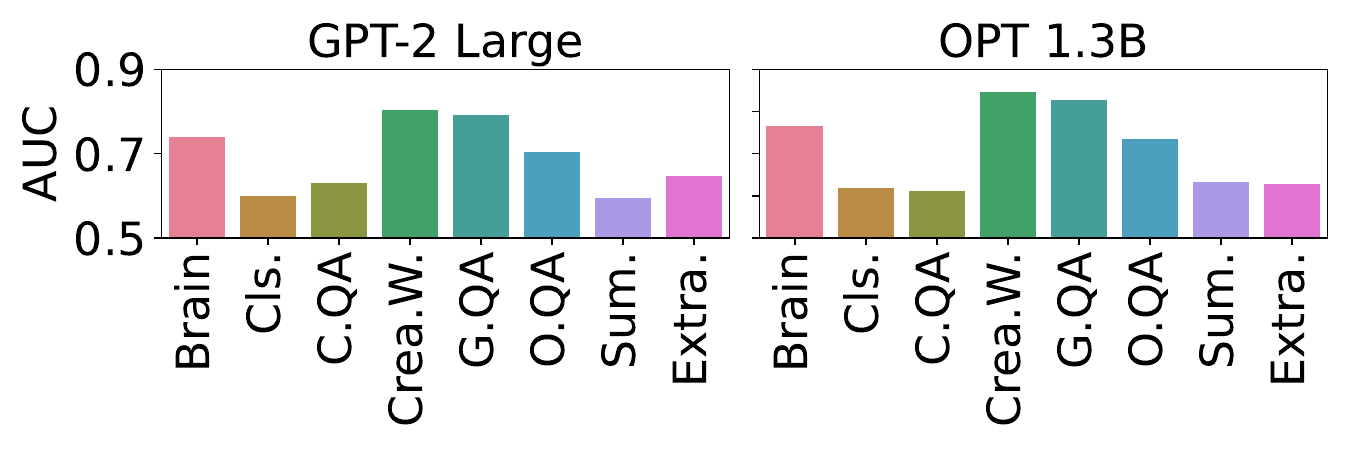}
\vspace{-20pt}
\caption{AUC score of MIAs per NLP task in Dolly dataset when attacking KD students. \textbf{The vulnerability to MIA varies across tasks}. KEYS-- Brain:Brainstorming; Cls.: Classification; C.QA: ClosedQA; G.QA: GeneralQA; O.QA: OpenQA; Sum.: Summarization; Extra.: Information Extraction. 
}

\label{fig:per_category}
\end{figure} 
We investigate the agreement between memorization and membership across eight NLP tasks.

To assess per-task membership, we sample both member and non-member data from the same NLP task and perform the \stable. 
Figure~\ref{fig:per_category} reports the average AUC across all KD techniques on GPT2-Large and OPT-1.3B. Our findings reveal that: i) Creative Writing, General QA, and BrainStorming exhibit the highest membership privacy risk, with AUC close to 0.90; while ii) Classification, Summarization, and ClosedQA demonstrate the lowest membership privacy risk, with AUC under 0.60. In contrast, we observe that classification and ClosedQA NLP tasks have the highest memorization. 
This discrepancy reveals an interesting phenomenon that \textbf{memorization is not membership in KD}, highlighting the need for designing more effective privacy attacks.

\section{Per-Block Privacy Risk Analysis}
\begin{figure}[h]
\centering
\includegraphics[width=1\linewidth]{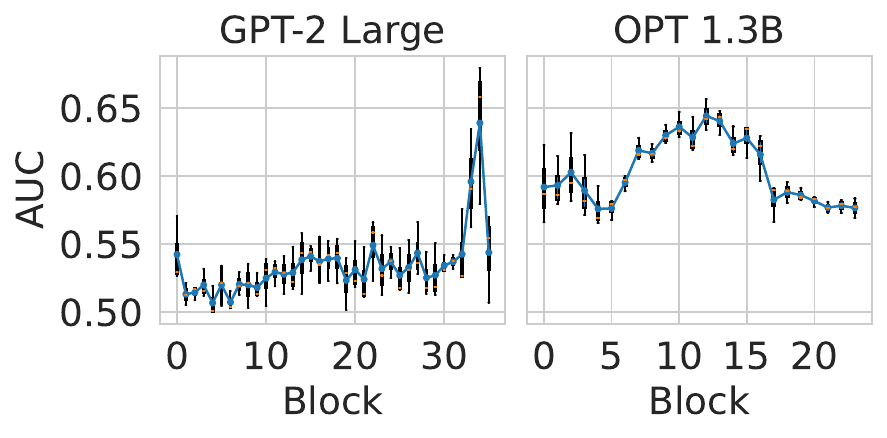}
\vspace{-15pt}
\caption{Block-wise privacy analysis. We compute the AUC score for inferring membership per each model block. \textbf{The privacy risk varies for different blocks.}}
\label{fig:Per-Layer-Analysis}
\end{figure} 

Our findings underscore the importance of carefully designing privacy-friendly KD strategies that selectively mitigate the high-risk components of the teacher model. One way to do that is through per-block privacy analysis to see how much each block relates to the privacy leakage. We design an analysis framework to quantify per-block privacy risk (details in App.~\ref{app:per-block}). Figure~\ref{fig:Per-Layer-Analysis} shows the per-block privacy leakage on GPT2-Large (left) and OPT-1.3B (right). Note that GPT2-Large has 36 blocks, and OPT-1.3B has 24 blocks. The x-axis is the block index, and the y-axis is the AUC score when only perturbing this block. We can observe that the privacy lekage of different blocks varies by a large margin. For example, on GPT2-Large, the 34-th and 33-th blocks are more vulnerable. The AUCs of the 33rd and 34th blocks are over 0.60 and 0.65, respectively. But for shallow blocks such as the 5-th block, the AUC is nearly 0.50.
Similarly, the most vulnerable blocks for OPT-1.3B have nearly 0.65 AUCs (the 12th and 13th blocks), but the AUCs of the 23rd and 5th blocks are below 0.60.
We can also observe that the vulnerable blocks are different for various models. For GPT2-Large, the deep blocks are more vulnerable. For OPT-1.3B, the middle blocks are more vulnerable.

\section{Conclusion}
\label{sec:discussion}

In this paper:
\begin{itemize}
    \item  we comprehensively study the empirical privacy protection (membership privacy and memorization) achieved by LLM KD techniques. 
    \item we also quantify privacy leakage per block and demonstrate that the vulnerabilities of different blocks are different.
    \item We identify and explain the privacy risks of LLM distillation, including the initialized student blocks from the teacher blocks, and the optimization process for KD loss function. 
\end{itemize}

\section{Limitation}

Although this paper has performed a comprehensive study on the privacy risk of KD and LLM blocks, it still has several limitations.
First, we focus on the empirical study of the privacy leakage and aim to perform a comprehensive evaluation. Theoretical analysis would be a potent supplement to this paper. We leave the theoretical analysis of the privacy leakage in KD as a future work.
Second, this paper doesn't provide defense solutions to mitigate privacy risks, such as selecting less vulnerable blocks to initialize the student model. We also leave the defense as part of the future work.

\bibliography{custom}

\newpage
\appendix

\section{Analysing Public Data versus Private Data}
\label{app:publicvsprivate}

\myparagraph{Distribution Shift Mitigation.}
Distribution shift between \pubdata and \pridata is one concern of rigorous MIA evaluation~\cite{duan2024membership}. According to~\cite{duan2024membership}, two important causes of distribution shift for LLM dataset are data source shift and temporary shift. Data source shift means the sources to collect member data and non-member data are different. Temporary shift means the time to generate text (e.g. arxiv papers) between member data and non-member data are different. The reason of the distribution shifts is that the member data and non-member data are not generated at the same time.

Our way to construct the dataset can mitigate both distribution and temporary shift. For the distribution shift, we randomly partition the Dolly dataset into two splits to guarantee the distribution consistency. Both \pubdata and \pridata are drawn from Dolly dataset and there is no shift. For the temporary shift, the Dolly dataset is generated during a short time window (March to April in 2023). \pubdata and \pridata are uniformly drawn from this window, thus the shift is mitigated.

\myparagraph{Data Leakage Measurement.}
One concern of evaluating MIA on LLM is the data leakage between \pubdata and \pridata~\cite{duan2024membership}. To mitigate this concern, we follow~\cite{duan2024membership} and measure the 7-gram similarity. The result indicates that the percentage of 7-gram similarity is 1.1\%. This low percentage suggests that the overlap between public and private data is minimal.

\section{Teacher Results of Low FPR}
\label{app:teach_low_fpr}

Table~\ref{tab:MIA_teachers_low_fpr} shows the MIA results of TPR at low FPR regions, including TPR@001 and TPR@0. We can observe that MIAs can reveal a large amount of teacher's membership privacy, and the most successful MIA differs across models and metrics.

\begin{table*}[t]
    \small
    \centering
\begin{tabular}{@{}llccccccc@{}}
\toprule
             &     &   \loss & \mink & \minkpp & \stable & \pretrain & \zlib & \mope \\ \midrule
\multirow{2}{*}{GPT-2 XL}      & TPR@001 & 0.0173 & 0.0168 & 0.0083 & 0.0538 & 0.0068 & 0.0229 & 0.0000\\
                                & TPR@0   & 0.0035 & 0.0005 & 0.0017 & 0.0203 & 0.0007 & 0.0139 & 0.0000\\

                                \midrule
\multirow{2}{*}{OPT 2.7B}     &    TPR@001 & 0.0237 & 0.0163 & 0.0095 & 0.0711 & 0.1993 & 0.0310 & 0.0137\\
                                & TPR@0   & 0.0047 & 0.0035 & 0.0045 & 0.0014 & 0.0244 & 0.0078 & 0.0000\\
                                \midrule
\multirow{2}{*}{LLAMA2-7B}     &    TPR@001 & 0.1001 & 0.0247 & 0.0025 & 0.2169 & 0.2966 & 0.2447 & 0.1005\\
                                & TPR@0   & 0.0486 & 0.0195 & 0.0002 & 0.0.0047 & 0.0062 & 0.0152 & 0.0503\\                                

\bottomrule
\end{tabular}
\caption{Membership privacy leakage of teachers (GPT2-XL, OPT-2.7B, and LLAMA2-7B) evaluated using TPR@001 and TPR@0 of seven MIAs once performing attack directly on teachers directly. Main takeaways: \textbf{i) All teachers exhibit significant membership privacy leakage, though the extend varies across families; and ii) The most successful MIA differs across families and metrics, highlighting the absence of a universally optimal MIA strategy.}}
\label{tab:MIA_teachers_low_fpr}
\end{table*}

\section{ROC curves}
\label{app:roc}

\begin{figure*}[t]
\centering
\includegraphics[width=\linewidth]{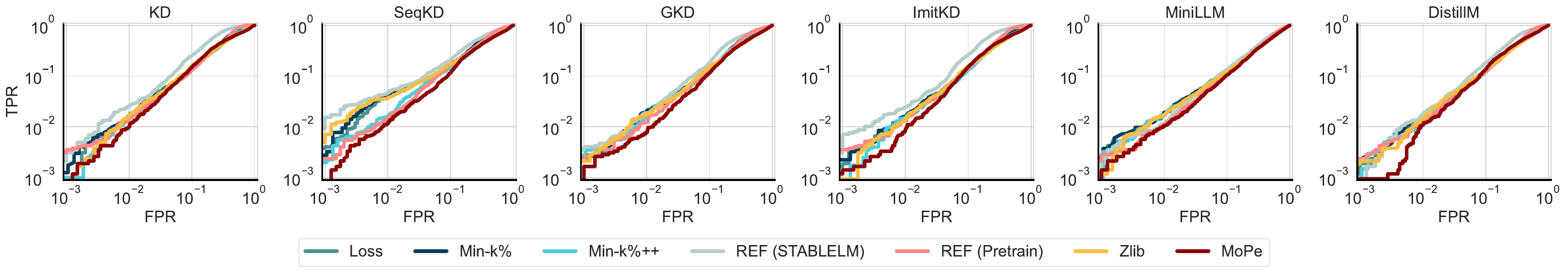}

\includegraphics[width=\linewidth]{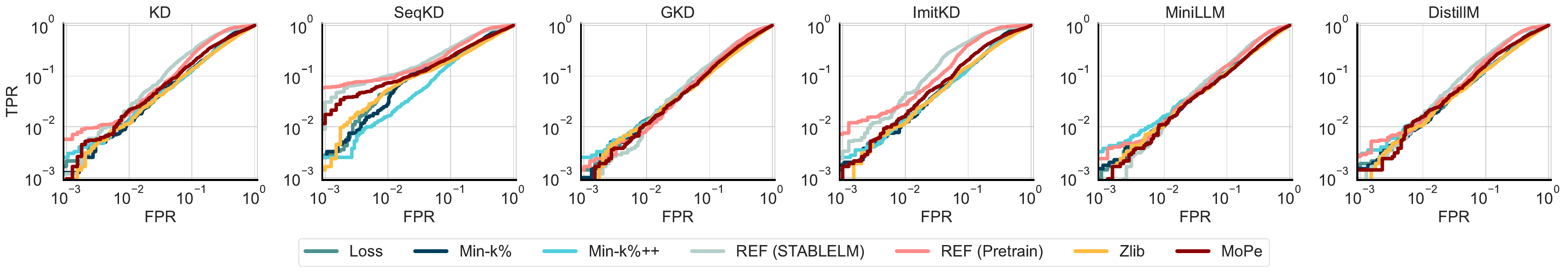}

\caption{Membership privacy protection of six knowledge distillation techniques on the Dolly dataset. We report True-Positive Rate (TPR) versus False-Positive Rate (FPR) of 7 membership inference attacks against GPT-2 Large (first row) and OPT-1.3B (second row). \textbf{\pretrain and \stable perform better at low FPR region.} This phenomenon is consistent with the observation of attacking the teachers.}
\label{fig:roc_curve}
\end{figure*} 

We also plot the log-scale ROC curve over the two large student models, GPT2-Large and OPT-1.3B. In Figure~\ref{fig:roc_curve}, the first row shows the curves of GPT2-Large and the second row shows OPT-1.3B. We can observe that \stable and \pretrain perform better at low FPR regions. This phenomenon aligns with the attack performance on \teacher (Table~\ref{tab:MIA_teachersGPTOPT}), where reference-based attacks have much higher TPR@05 and TPR@01 than other attacks.

\section{TPR at Low FPR regions}
\label{app:tpr}
\begin{figure*}[h]
\centering
\includegraphics[width=\linewidth]{kd_figs/gpt2/heatmap_AUC.pdf}

\includegraphics[width=\linewidth]{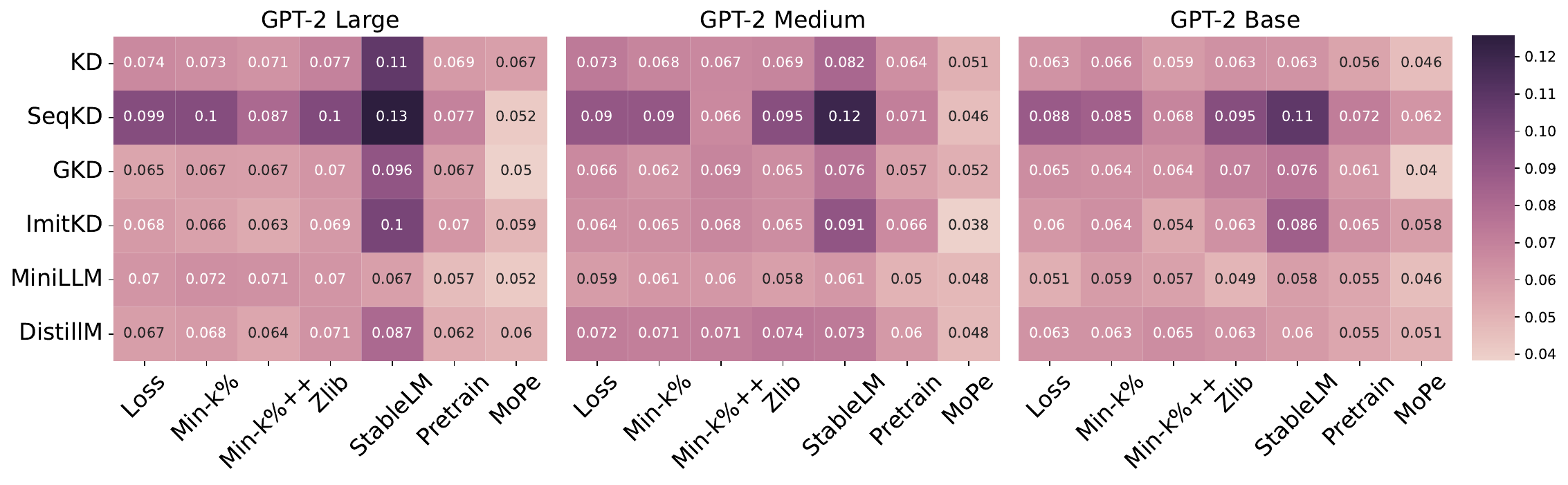}

\includegraphics[width=\linewidth]{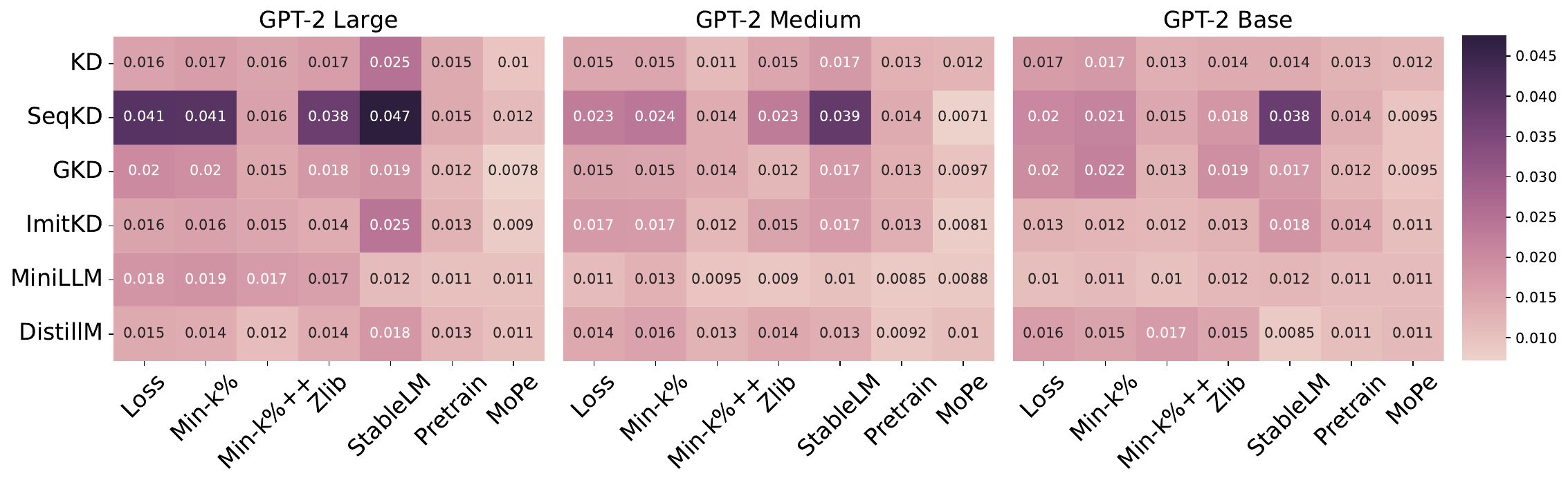}

\caption{Membership privacy protection of six knowledge distillation techniques using GPT-2 family. We report the AUC score of seven MIAs at inferring private training members of teachers based on three students (GPT-2 Large, GPT Medium and GPT Small) obtained through various knowledge distillation techniques. \textbf{None of knowledge distillation techniques can create students that protect privacy of their GPT-2 XL teacher.}}
\label{fig:model_AUC_GPT2_studnets_TPR}
\end{figure*}

\begin{figure*}[h]
\centering
\includegraphics[width=\linewidth]{kd_figs/opt/heatmap_AUC.pdf}

\includegraphics[width=\linewidth]{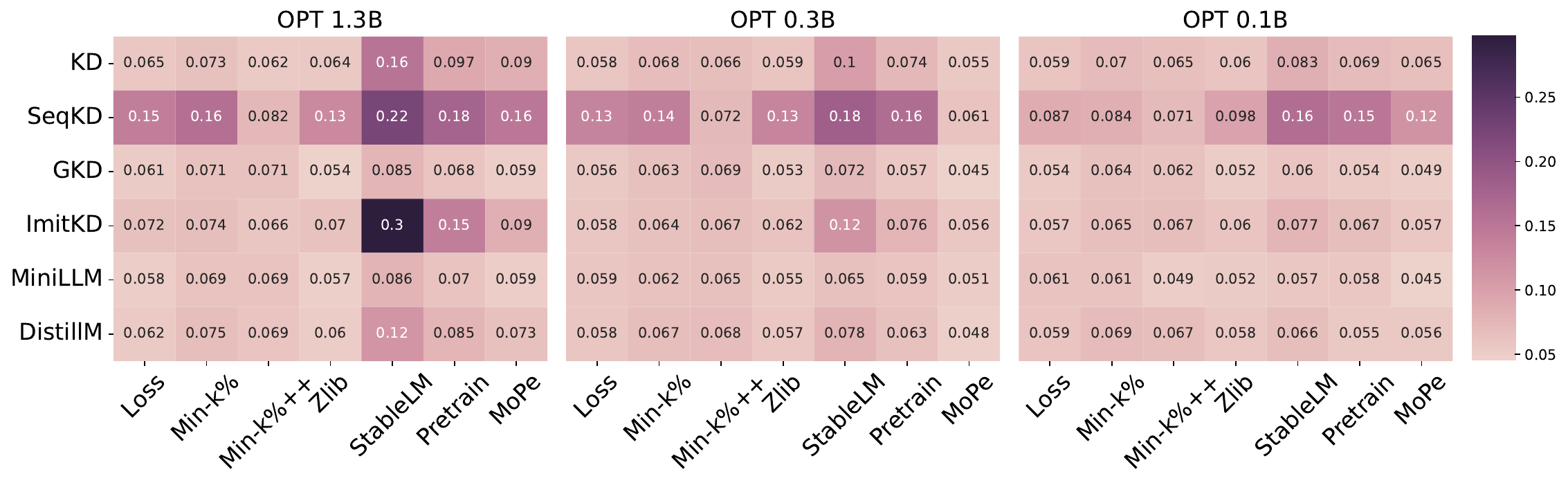}

\includegraphics[width=\linewidth]{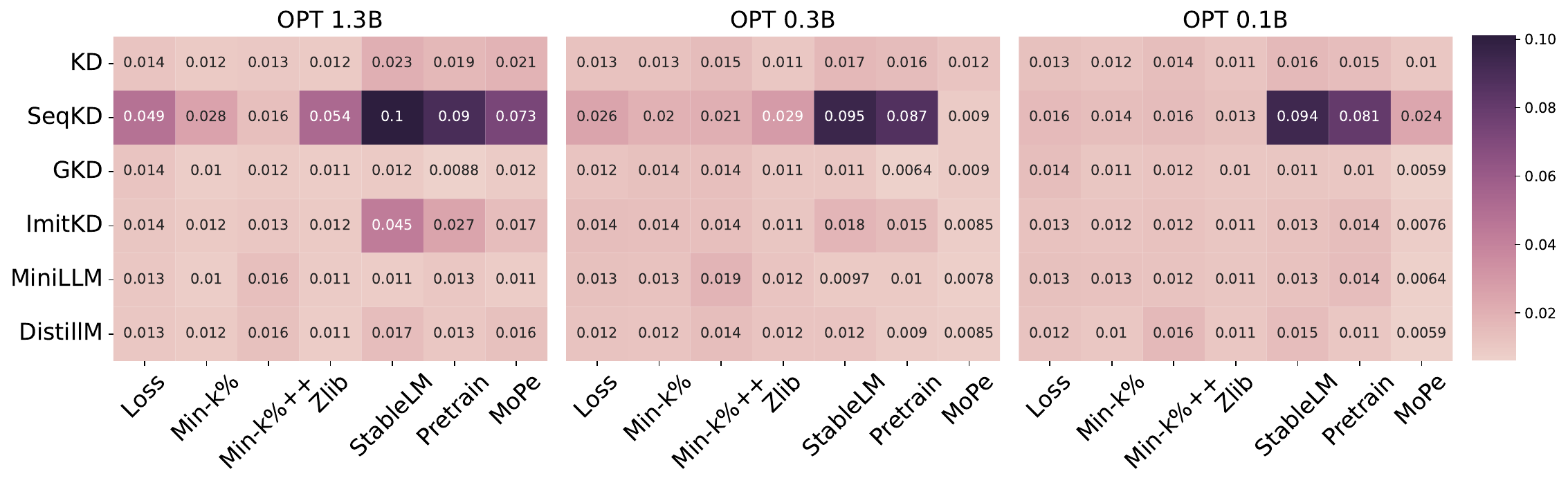}
\caption{Membership privacy protection of six knowledge distillation techniques using OPT family. We report the AUC score (first row), TPR at 5\% FPR (second row) and 1\% FPR (last row)  of seven MIAs at inferring private training members of teachers based on three students (OPT 1.3B, 0.3B and 0.1B) obtained through various knowledge distillation techniques. \textbf{None of knowledge distillation techniques can create students that protect privacy of their OPT teacher.}}
\label{fig:model_AUC_OPT_studnets_TPR}
\end{figure*}

Figure~\ref{fig:model_AUC_GPT2_studnets_TPR} and Figure~\ref{fig:model_AUC_OPT_studnets_TPR} shows the comprehensive results of attacking the student models. In both figures, the first, second, and third row shows the results of AUC, TPR@05, and TPR@01, respectively. The observations of TPR@05 and TPR@01 are similar to the results of AUC. None of the KD techniques can create a student that protect the privacy of the teacher. And two reference-based attacks (\stable and \pretrain) achieves the highest attack performance.

\section{Utility}
\label{app:utility}

\begin{figure*}[h]
\centering
\includegraphics[width=0.45\linewidth]{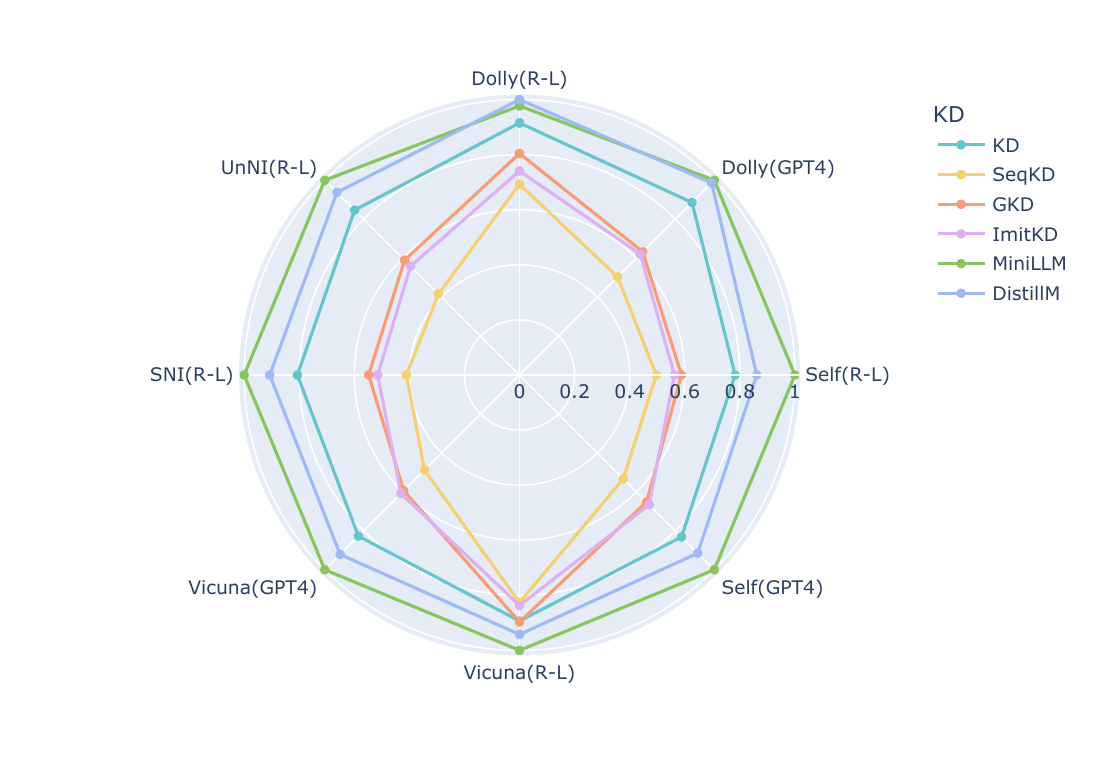}
\includegraphics[width=0.45\linewidth]{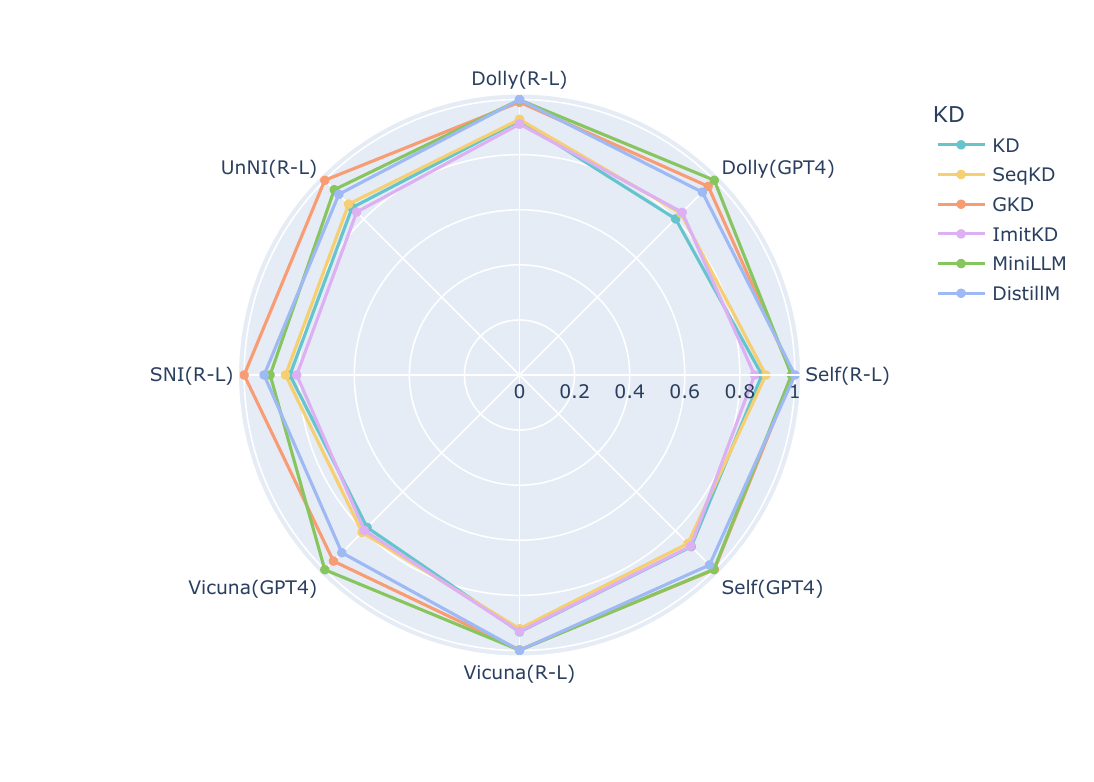}
\caption{Utility evaluation of GPT2-Large (left) and OPT-1.3B (right). We follow~\cite{ko2024distillm} to report the performance over five datasets.}
\label{fig:utility}
\end{figure*}

Figure~\ref{fig:utility} shows the utility score of GPT2-Large and OPT-1.3B. We follow~\cite{ko2024distillm} to report the utility scores over five datasets: Dolly test set~\cite{DatabricksBlog2023DollyV2}, Self-Instruct~\cite{wang2022self}, Vicuna evaluation, Super-Natural Instruction~\cite{wang2022super}, and Unatural Instruction~\cite{honovich2022unnatural}. Following LLM knowledge distillation literature~\cite{ko2024distillm,gu2023minillm}, we quantify the utility of teachers and students using Rouge-L score~\cite{lin2004rouge} and GPT-evaluated score. The higher the Rouge-L and GPT-evaluated score, the better the utility.

\section{LLAMA2-7B}
\label{app:llama}
\begin{figure*}[h]
\centering
\includegraphics[width=0.3\linewidth]{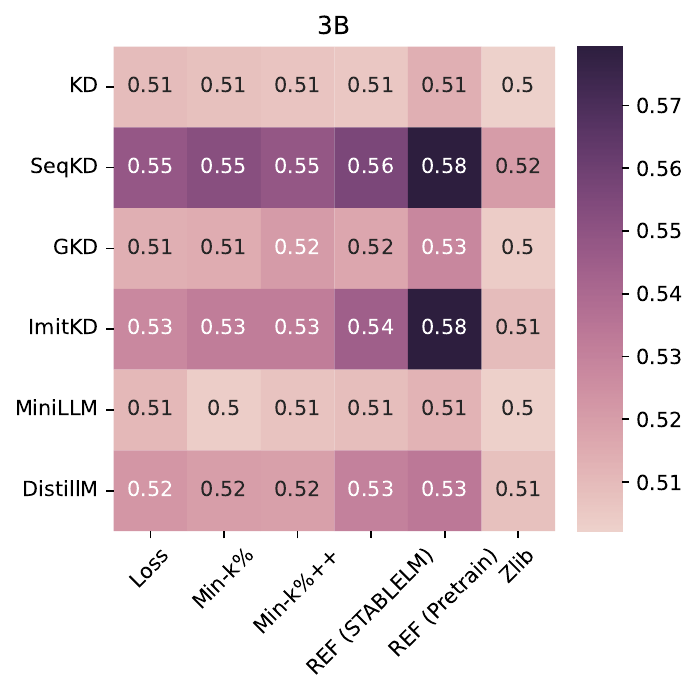}
\includegraphics[width=0.3\linewidth]{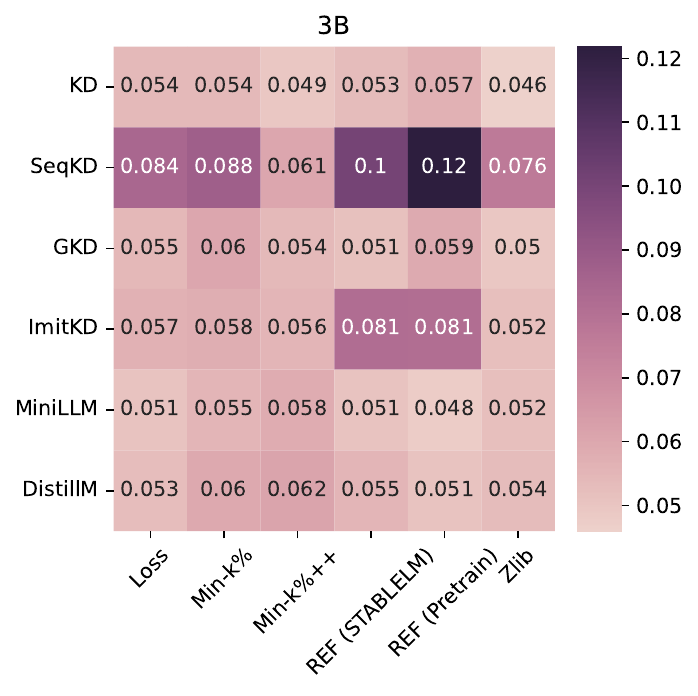}
\includegraphics[width=0.3\linewidth]{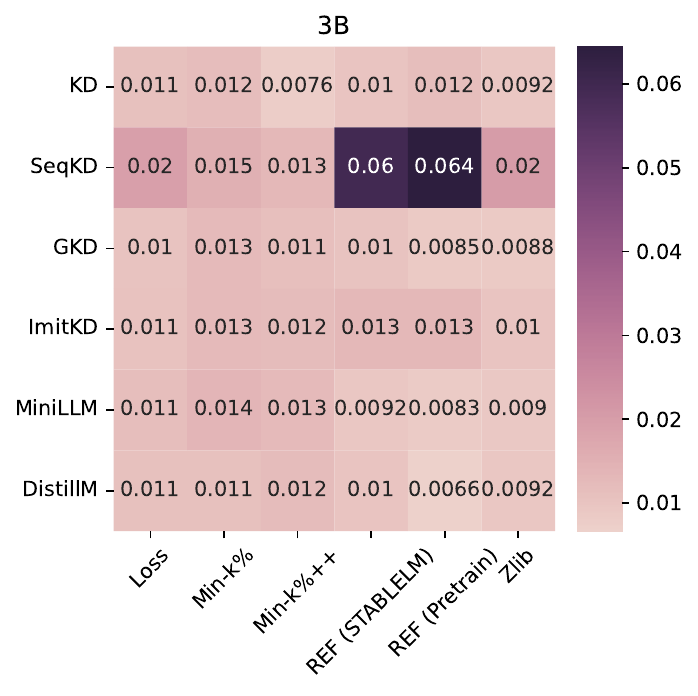}
\caption{Membership privacy protection of six knowledge distillation techniques using LLAMA2-7B. We report the AUC score (left column), TPR at 5\% FPR (middle column) and 1\% FPR (right column) of seven MIAs at inferring private training members of teachers based on LLAMA2-3B student obtained through various knowledge distillation techniques. \textbf{None of knowledge distillation techniques can create students that protect privacy of their LLAMA2-7B teacher.}}
\label{fig:model_AUC_LLAMA_studnets}
\end{figure*}

Figure~\ref{fig:model_AUC_LLAMA_studnets} shows the AUC, TPR@05, and TPR@01 on LLAMA-0.3B. LLAMA-3B can still some of teacher's privacy.

\section{Per-Block Analysis Algorithm}
\label{app:per-block}

Our framework perturbs only one block and use the model's output perturbation to measure the block sensitivity  by leveraging MoPe~\cite{li2023mope}. The pipeline is depicted in Algorithm~\ref{alg:per-block}. We first break the model into multiple blocks. For each block, we isolate it from other blocks and add noises to the block weights to generate multiple perturbed blocks. We combine the perturbed block with other isolated blocks to construct perturbed models. For each perturbed model, we compute the CE loss over all samples in $\mathcal{D}_{\text{mem}}$ and $\mathcal{D}_{\text{non}}$ and compute the average loss deviation $\Delta l_{i,\text{mem}}$ and $\Delta l_{i,\text{non}}$. We use the deviation to evaluate MIA performance and use the AUC score as an index of membership privacy leakage.

\begin{algorithm}[tb]
\SetKwComment{Comment}{{\footnotesize$\triangleright$\ }}{}
   \caption{Per-block privacy analysis framework.}
   \label{alg:per-block}
   \KwIn{A target model $\mathcal{M}$ containing {$ \{\theta^l\}_{l=1}^L$} blocks, A set of member data $\mathcal{D}_{\text{mem}}$, A set of non-member data $\mathcal{D}_{\text{non}}$, the number of perturbed models $N$.}
   \KwOut{Privacy leakage of each block}
   \BlankLine
\begin{algorithmic}[1]

\FOR {$\mathbf{x}_{\text{mem}} \in \mathcal{D}_{\text{mem}}$}
\STATE$\text{Loss}_{\text{mem}}=\mathcal{M}(\mathbf{x}_{\text{mem}})$\Comment*[r]{{\scriptsize Compute member loss}}
\ENDFOR
\FOR {$\mathbf{x}_{\text{non}} \in \mathcal{D}_{\text{non}}$}
\STATE$\text{Loss}_{\text{non}}=\mathcal{M}(\mathbf{x}_{\text{non}})$\Comment*[r]{{\scriptsize Compute non-member loss}}
\ENDFOR

\FOR{$l \in L$}
    \FOR{$n \in N$}
        \STATE $\theta^{'l}=\theta^l+\text{Noise}_n$ \Comment*[r]{{\scriptsize Perturb the block}}
        \STATE $\mathcal{M}_{l,n}=\{ \theta^1,...,\theta^{'l},...,\theta^{L} \}$ \Comment*[r]{{\scriptsize Perturbed model }}
        \FOR {$\mathbf{x}_{\text{mem}} \in \mathcal{D}_{\text{mem}}$}
        \STATE $\Delta \text{Loss}_{l,n,\text{mem}}=| \mathcal{M}_{l,n}(\mathbf{x}_{\text{mem}}) - \text{Loss}_{\text{mem}} |$ 
        \ENDFOR
        \FOR {$\mathbf{x}_{\text{non}} \in \mathcal{D}_{\text{non}}$}
        \STATE $\Delta \text{Loss}_{l,n,\text{non}}=| \mathcal{M}_{l,n}(\mathbf{x}_{\text{non}}) - \text{Loss}_{\text{non}} |$
        \ENDFOR
    \ENDFOR
    \STATE $\Delta \text{Loss}_{l,\text{mem}} =  \text{mean}_{n} (\Delta \text{Loss}_{l,n,\text{mem}})$ \Comment*[r]{{\scriptsize Average over $N$ perturbations to compute average variance}}
    \STATE $\Delta \text{Loss}_{l,\text{non}} = \text{mean}_{n} (\Delta \text{Loss}_{l,n,\text{non}})$ \Comment*[r]{{\scriptsize Average over $N$ perturbations to compute average variance}}

    \STATE Use membership signals $\Delta \text{Loss}_{l,\text{mem}}$ and $\Delta \text{Loss}_{l,\text{non}}$ to compute per-block privacy leakage, $\text{AUC}_l$
\ENDFOR
\STATE Return $\{\text{AUC}_l\}_{l=1}^L$

\end{algorithmic}
\end{algorithm}

\end{document}